\documentclass{article}
\usepackage{spconf,amsmath,graphicx}
\usepackage{epstopdf}
\usepackage{flushend,color}

\usepackage{fancyheadings}
\title{Generating Binary Tags for Fast Medical Image Retrieval \\ Based on Convolutional Nets and Radon Transform}
\name{Xinran Liu$^\dagger$, H.R.Tizhoosh$^\ddagger$, J.Kofman$^\dagger\ddagger$}
\address{$^\dagger$ Department of Systems Design Engineering\\
University of Waterloo, Waterloo, ON, Canada N2L 3G1\\ \vspace{0.04in}
\texttt{x435liu@uwaterloo.ca}, \texttt{jkofman@uwaterloo.ca} \\ 
$^\ddagger$ Centre for Bioengineering and Biotechnology\\
University of Waterloo, Waterloo, ON, Canada N2L 3G1\\
\texttt{tizhoosh@uwaterloo.ca}}
\begin{document}
\maketitle
\begin{abstract}
Content-based image retrieval (CBIR) in large medical image archives is a challenging and necessary task. Generally, different feature extraction methods are used to assign expressive and invariant features to each image such that the search for similar images comes down to feature classification and/or matching. The present work introduces a new image retrieval method for medical applications that employs a convolutional neural network (CNN) with recently introduced Radon barcodes. We combine neural codes for global classification with Radon barcodes for the final retrieval. We also examine image search based on regions of interest (ROI) matching after image retrieval. The IRMA dataset with more than 14,000 x-rays images is used to evaluate the performance of our method. Experimental results show that our approach is superior to many published works.
\end{abstract}

\section{Introduction}
For more than two decades, content-based image retrieval (CBIR) has been a very active research discipline \cite{akgul2011content,smeulders2000content}. As a branch of computer vision, CBIR aims to search for digital images in large databases or archives based on the \lq\lq content\rq\rq of images, such as colors, shapes, textures or any other information that can be derived from the image itself. In a generic CBIR system, given a user-supplied query image, the system is supposed to search the database and return the images that have high similarity to the user's query image. For medical applications, such a system could assist clinicians in more reliable diagnosis or timely detection of malignancies by retrieving similar cases from the image archive or database. Existing computer-aided diagnosis/detection systems are mainly focused on specific areas, such as breast masses, lung nodules and colonic polyps, which target very specific image features without using any comparison of the current case with existing cases through image search \cite{akgul2011content}.

Medical images have unique characteristics that make them different from other images of our daily life. For instance, most medical images are gray-level (no color). Another attribute of medical images is that images captured from the same body region (head, thorax, abdomen etc.), to a great extent, exhibit large similarities, even from different individuals. Furthermore, the most valuable information in medical images usually appears to be located in a very small image region, known as the region of interest (ROI), which in most cases, represents a type of abnormality or malignancy. Not only global properties but also specific local features thus should be considered when a medical image retrieval system needs to be designed. 

In this paper, a short binary code named convolutional neural network code, short CNNC, is generated based on a well-trained deep convolutional neural network and then used to annotate a medical image. A new method for medical image retrieval is proposed based on CNNC and Radon Barcode (RBC). ROI matching is then implemented for the retrieved images. The remainder of the paper is organized as follows: In Section \ref{sec:CBIR}, we give a brief review of CBIR research. In Section \ref{sec:prop}, we describe our approach for medical image retrieval and ROI matching, respectively. In Section \ref{sec:exp}, we report qualitative and quantitative experimental results. Section \ref{sec:conc} concludes the paper.

\section{Brief Review of CBIR Literature} 
\label{sec:CBIR}
Inspired by the methods in general CBIR systems, some achievements have been made by permeating these methods into medical applications. Wei et al. proposed a Gabor filtering method to extract the textural features for mammogram retrieval \cite{wei2007effective}. Greenspan et al. used multi-dimensional feature space to represent the image and extract coherent regions by using unsupervised clustering via Gaussian mixture modeling, and then match images via the Kullback–Leibler measure \cite{greenspan2007medical}. Lehmann et al. compared the performance of different approaches for automatic categorization of medical images \cite{lehmann2005automatic}. Tommasi et al. developed a multi-cue approach based on the support vector machine (SVM) algorithm to annotate medical images automatically by combining global and local features \cite{tommasi2008discriminative} \cite{tommasi2009svm}, and achieved very good results in the ImageCLEF 2009 medical image annotation task \cite{tommasi2010overview}. Dimitrovski et al. proposed a hierarchical multi-label classification (HMC) system for medical image annotation \cite{dimitrovski2011hierarchical}. Fushman et al. employed a supervised machine learning approach by associating text-based and content-based image information to retrieve clinically relevant images \cite{demner2009annotation}. Ayed proposed a new descriptor for radiological image retrieval based on fuzzy shape contexts, Fourier Transforms and eigenshapes \cite{ben2012rotation}. Camlica et al. have proposed both autoencoders and SVM for medical images retrieval \cite{camlica2015autoencoding, camlica2015medical}. 

Recent annotation methods of medical images seem to be moving toward binary descriptors, away from high-dimensional feature vectors. This is mainly because binary codes are computationally faster and require less space, making image retrieval feasible to be applied on \lq\lq big data\rq\rq. Calonder et al. used binary strings and developed an efficient feature point descriptor called \lq\lq BRIEF\rq\rq \cite{calonder2010brief}. Rublee et al. proposed a new descriptor called \lq\lq ORB\rq\rq based on BRIEF, which is rotation invariant and resistant to noise \cite{rublee2011orb}. Leutenegger et al. proposed a novel method, named Binary Robust Invariant Scalable Keypoints (BRISK) for image key point detection, description and matching \cite{leutenegger2011brisk}, and achieved high-quality performance at a dramatically lower computational cost. In 2015, RBC based on the radon transform was used by Tizhoosh \cite{tizhoosh2015barcode}, and achieved good overall results for medical image retrieval on the IRMA dataset. What is lacking in the literature is a learning framework applied on specific domains (such as medical imaging) and validated using large publicly available datasets. 

Recently, artificial deep neural networks have been gaining increased use, since the method of training a deep neural network was proposed by Hinton in 2006 \cite{hinton2006reducing}. Krizhevsky et al. first applied a deep convolutional neural network to the ImageNet LSVRC-2010 dataset and achieved the best result in that challenge, showing that the deep neural network is a powerful tool for image classification and retrieval \cite{krizhevsky2012imagenet}. However, rigorous validation for image retrieval with publicly available benchmark datasets such as IRMA images seems to be missing. As well, as the training requirement limits the usability of deep networks (because most data are distributed and cannot be gathered in a central computing node with GPUs for training), one may be inclined to rather combine deep representations with non-learning binary descriptors to fuse the strengths of both technologies. 

In this work, we combine convolutional neural network codes (CNNCs) with recently introduced Radon barcodes (RBCs), where the former handles the task of finding the top 50 hits/matches, and the latter is employed for more refined search. As well, we conducted preliminary work on ROI matching. We used the IRMA dataset with more than 14,000 x-ray images to validate our method.

\section{Method}
\label{sec:prop}
\subsection{Convolutional Neural Network Code (CNNC)}
A convolutional neural network (CNN) is a type of feed-forward artificial neural network that is designed to use minimal image pre-processing and work on raw image data. After proper training, the CNN is usually able to extract some features from the training set and predict what category it belongs to when a new input image is encountered. However, the parameters in convolutional kernels used for feature extraction are totally learned automatically by the neural network itself, which is different from other (purposefully designed) algorithms for feature extraction based on common computer vision methods. Although a very deep convolutional neural network has been shown to have better performance on image classification \cite{simonyan2014very}, we use a relatively shallow architecture mainly for the sake of time efficiency and limited computational power.

\begin{figure*}[tb]
\centering
\vspace{0.05in}
\centerline{\includegraphics[width=0.9\textwidth]{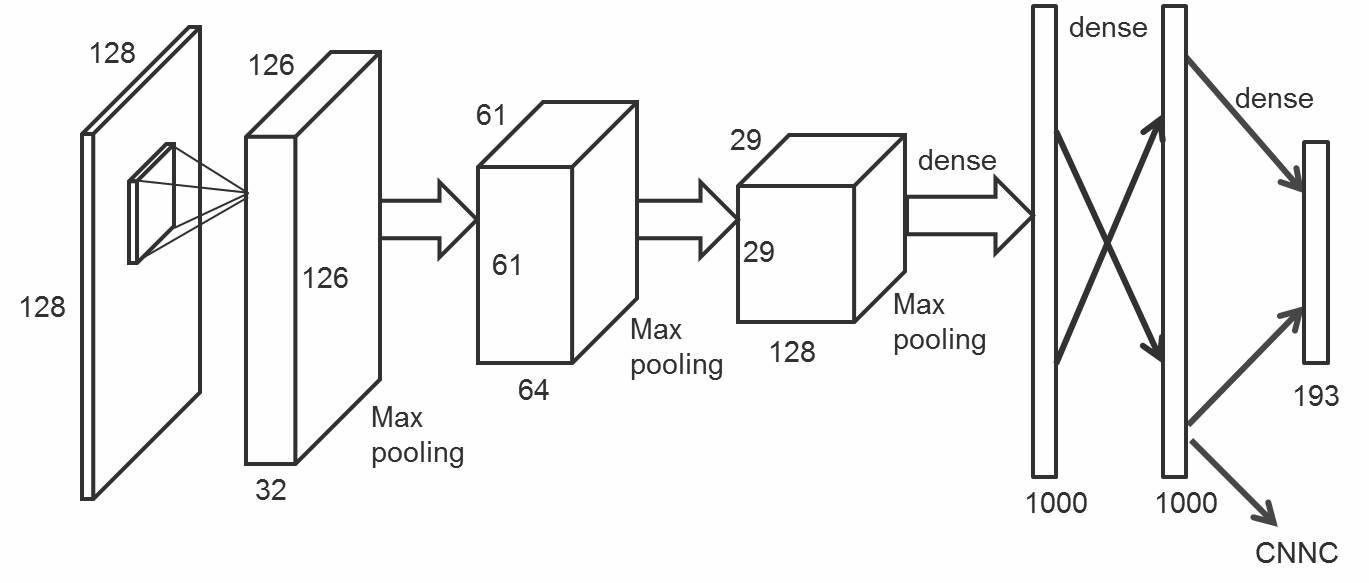}}
\caption{Illustration of the architecture of the proposed convolutional neural network.}
\label{fig:CNN}
\end{figure*}

\begin {table}[tb]
\caption{Architecture settings for CNN}
\label{tab:architect}
\begin{center}
\begin{tabular}{|c|c|c|c|c|}
\hline
Layer & Stage & channels & Kernel & Pooling \\
\hline
1 & conv+max & 32 & 3$\times$3 & 2$\times$2 \\
\hline
2 & conv+max & 64 & 3$\times$3 & 2$\times$2 \\
\hline
3 & conv+max & 128 & 3$\times$3 & 2$\times$2 \\
\hline
4 & full & 1000 & - & - \\
\hline
5 & full & 1000 & - & - \\
\hline
6 & full & 193 & - & - \\
\hline
\end{tabular}
\end{center}
\end{table}

As shown in Figure \ref{fig:CNN}, the CNN we use consists of five layers; the first three are convolutional and the remaining two are fully-connected layers. The output of the last fully-connected layer is fed to a 193-way softmax unit. All images in the IRMA dataset are down-sampled to a fixed resolution of 128$\times$128. The Rectified Linear Units (ReLU) \cite{glorot2011deep} is applied to the output of all five layers since it can speed up the training procedure several times faster than using the traditional sigmoid or \emph{tanh} functions. The non-overlapping max-pooling is applied after each convolutional layer to reduce the number of parameters in the CNN. We detail the architecture size in Table \ref{tab:architect}.

To prevent the neural network from overfitting, dropout \cite{srivastava2014dropout} is used in the first two fully-connected layers, where the output of each hidden neuron is set to zero with probability 0.5. In this way, the neural network will sample a different architecture when an input image is presented, benefitting the generalization of the neural network. Data augmentation is another technique that we use to prevent overfitting, which enlarges the training set by randomly shifting the image both horizontally and vertically within a small range. 

After training, the neural network can predict the class label when a test image is presented. For the purpose of image retrieval, we consider the feature activations at the last, 1000-dimensional fully-connected layer and generate an array of outputs of all neurons in that layer. Note that because the ReLU and Dropout are used in this layer, we use the neuron output before ReLU and Dropout. Then the Convolutional Neural Network Code (CNNC) can be obtained after binarization by selecting a threshold value zero.

\subsection{Radon Barcodes (RBCs)}

\begin{figure}[tb]
\centering
\centerline{\includegraphics[width=0.9\columnwidth]{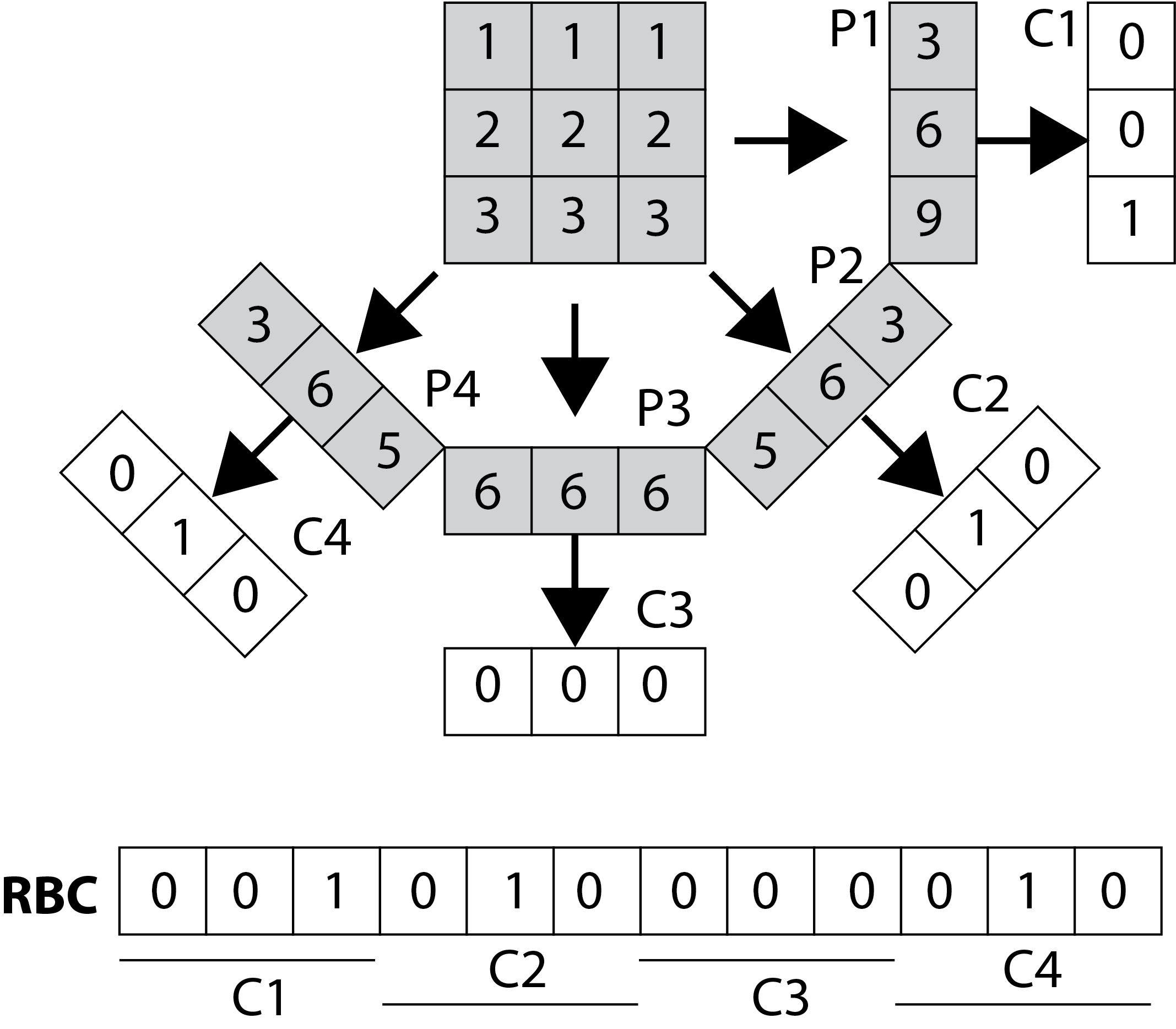}}
\caption{Radon barcodes according to Tizhoosh \cite{tizhoosh2015barcode}.}
\label{fig:RBC}
\end{figure}

The Radon barcode annotation is inspired by the Radon Transform. RBCs have been meanwhile used for ROI location in breast ultrasound images \cite{Tizhoosh2016}. The generation of RBC is detailed in \cite{tizhoosh2015barcode}.  First, the image is down-sampled to a fixed resolution. Second, the Radon Transform is applied to get the projection along a specific projection angle. Third, different projections are obtained by changing the projection angle and then binarized based on a \lq\lq local\rq\rq threshold, generating a code fragment. Last, all code fragments are connected to generate RBC of this image (Figure \ref{fig:RBC}).

The length of the RBC is determined by the size of the input image and the number of projection angles used in the Radon Transform. For larger images, more projection angles were applied, resulting in longer RBCs, thus, more information about the image can be described in the code. A very long RBC can represent the image more precisely since more projections can contribute to the accurate reconstruction of the image when the Radon transform is used. However, it could be more time-consuming if we use very long RBCs to annotate and retrieve all images in the dataset. Therefore, we only generate a long RBC on a sample of images after using CNNC to find a small subset of image candidates. In this paper, we down-sample the size of candidate images to 192$\times$192 and use 16 projection angles to generate the RBC. Note that for both CNNC and RBC, if two images produce the code with a small Hamming separation, it can be concluded that the two images have high similarity in their contents. 
\subsection{Image Retrieval Algorithm}
After training the convolutional neural network, all images in the dataset are again fed to the network to generate the CNNC. When a query image is presented, the CNNC and RBC are both generated. Then we compare the CNNC of the query image with all the images in the dataset and select 50 candidates that are similar to the query/input image (their codes have short Hamming distances to the input image). The long RBC is calculated for each of the top-50 matches/hits. We then compare the RBC of the query image with these candidates and re-rank them. At last, the top-10 candidates after re-ordering are returned to the user as the retrieval result. The error is calculated for the first hit only (the most similar image). We further implement ROI matching, which is described in the next section.

\subsection{ROI Matching}
\label{sec:ROImatch}
For medical applications, the ROI is the most important and useful region in the image, such as a tumor. For a given ROI in the query image, the aim of ROI matching is to find similar ROIs in all retrieved images. For this purpose, 2D cross-correlation is applied in this paper, since it is a simple method to measure the similarity of two images. Cross-correlation of 2D signals, also known as the sliding dot product, computes the element-by-element products of two matrices and sums them. 

For the query image and each of the 10 retrieved images, we first subtract the mean intensity of the image from each pixel. Then 2D cross-correlation is implemented by calculating the sliding dot product of the modified (mean-subtracted) retrieved images and the ROI from the modified query image. The maximum of the cross-correlation corresponds to the estimated most similar region in the retrieved image.

\section{Experiments and Discussion}
\label{sec:exp}
\subsection{IRMA Dataset}
\begin{figure*}[tb]
\centering
\vspace{0.05in}
\centerline{\includegraphics[width=13cm]{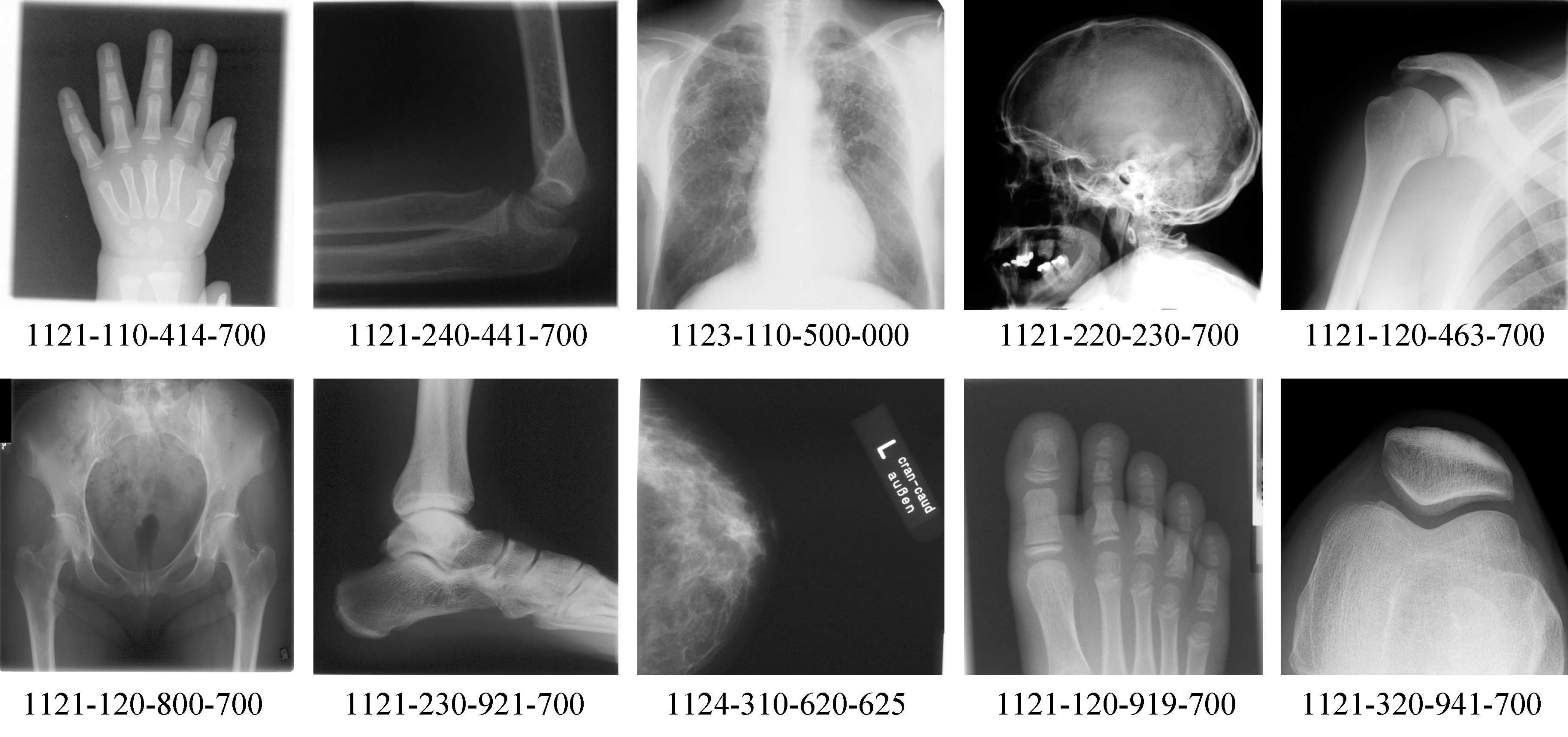}}
\caption{Sample images in the IRMA dataset with their corresponding IRMA codes for benchmarking.}
\label{fig:IRMAsamples}
\end{figure*}

The IRMA dataset is a collection of 14410 anonymized radiographs over 193 categories provided by the Department of Diagnostic Radiology, Aachen University of Technology (Fig. \ref{fig:IRMAsamples}) \cite{Lehmann2003,Lehmann2006}. Among those images, 12677 are training images and 1733 are for testing. All images were classified according to the IRMA code, which is a string of 13 characters and consists of four mono-hierarchical axes: the technical code T (imaging modality); directional code D (body orientations); anatomical code A (the body region); and biological code B (the biological system examined) \cite{tommasi2010overview}. Figure \ref{fig:IRMAsamples} shows 10 sample images in the IRMA dataset with their corresponding IRMA code.

\subsection{Image Retrieval Result}
\begin{figure*}[htb]
\centering
\vspace{0.05in}
\centerline{\includegraphics[width=0.83\textwidth]{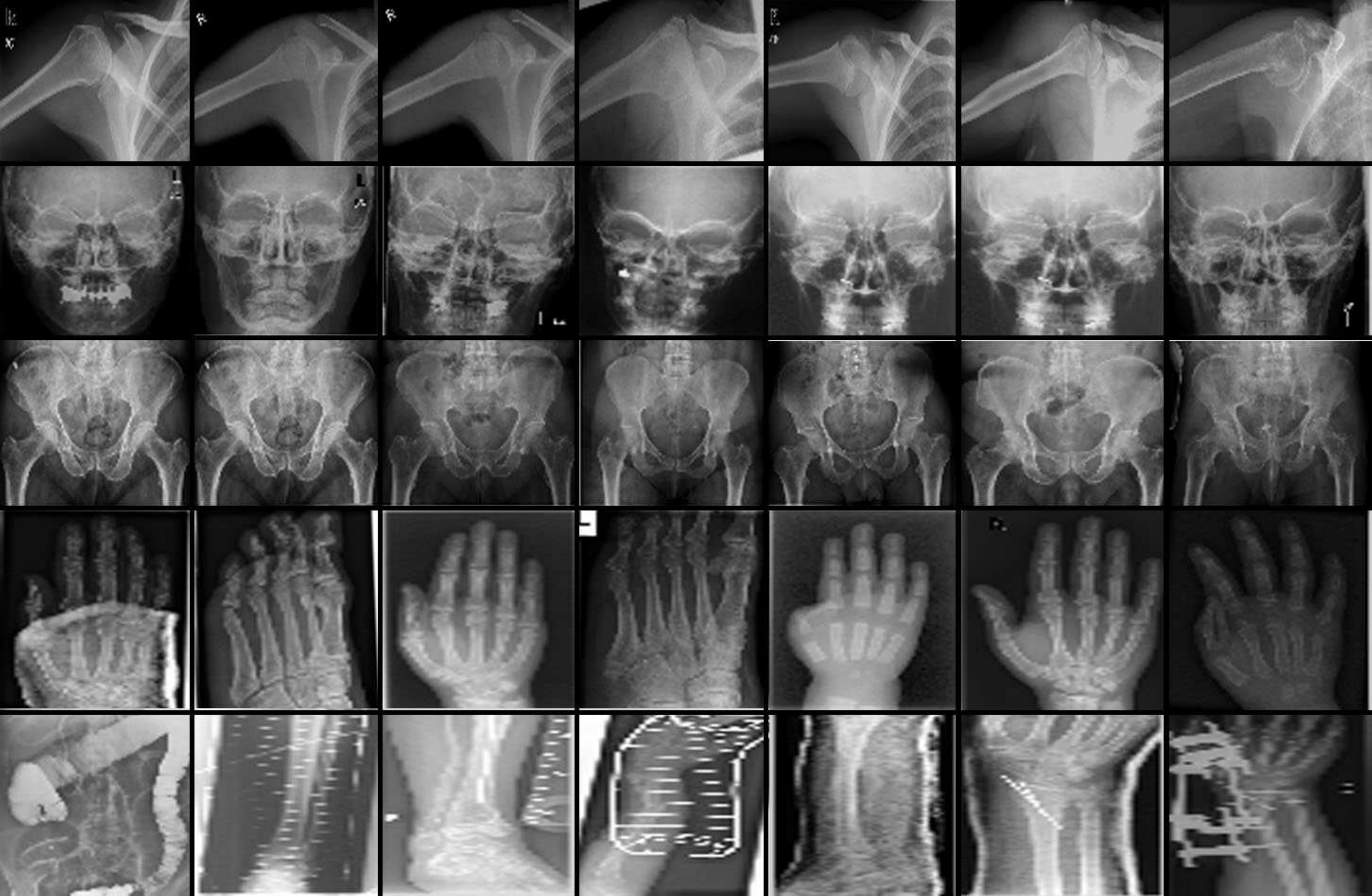}}
\caption{Five examples of image retrieval using the proposed method.}
\label{fig:RetreivalSamples}
\end{figure*}

The convolutional neural network was trained for 50 epochs, which took 5 hours on a NVIDIA Tesla C2050 GPU. All 1733 test images in the IRMA dataset were tested as query images by using the image retrieval algorithm described in this paper. Figure \ref{fig:RetreivalSamples} shows five image retrieval examples. The query image is in the first column while the remaining images are the images from the training set that have the highest similarity with the query image based on our method. For most X-ray images (top three rows in  Fig. \ref{fig:RetreivalSamples}), we were able to find the most similar images easily and accurately (accuracy can be calculated using IRMA codes). For some images (bottom two rows in  Fig. \ref{fig:RetreivalSamples}), the method was limited because a sufficient number of training images in the dataset was not avaiable for some classes. Generally, the IRMA dataset is challenging because of its imbalance; some image classes have hundreds of examples, whereas some others only have a few. Even 35\% of the classes have less than 10 training images. That makes learning difficult, but it may occur in real practice when the training set available may be small and therefore unbalanced. 

The error evaluation method of \cite{tizhoosh2015barcode} was used. The total error score over 1,733 test images is defined as follows:

\begin{equation}
E_{total}\left ( l^{query} \right )=\sum_{i=1}^{1733}\sum_{k=1}^{4}\sum_{j=1}^{n_{d}}\frac{1}{B_{j}^{ik}}\frac{1}{j}\delta \left ( l_{j}^{k,query}, l_{j} \right )
\end{equation}
where $n_{d}\in \left \{ 3, 4 \right \}$, $B_{j}^{ik}$ is the number of possible labels for position $j$. The function $\delta ( l_{j}^{k,query}, l_{j} )$ delivers 0 if $l_{j}^{k,query}=l_{j}$, otherwise 1.

Table \ref{tab:res} shows the resulting errors by comparing the method proposed in this paper with other published results \cite{tommasi2010overview} \cite{haas2012superpixel}. The total error score is 224.13 using images with a resolution of 128$\times$128 as the input to the CNN while the state-of-the-art error score on this dataset is 169.5 \cite{avni2009addressing}. However, our method is simpler than \cite{avni2009addressing} since little pre-processing is needed (their model used an additional label setting for fine-tuning). Furthermore, we reduced the resolution of images fed to the CNN to 96$\times$96, which increased the error score by 13.8. We also evaluated the error by using a CNNC without binarization. We expected that this approach could decrease the error score since it is a more accurate and detailed code. However, the error score increased by 32.19 when binarization was disabled.

\begin{table}[tb]
\caption{The retrieval error of the three approaches tested in this paper (bold) and some other published approaches from literature. A low score indicates a higher accuracy.}
\label{tab:res}
\begin{center}
\begin{tabular}{|l|c|}
\hline
Approach & Error Score \\
\hline
TAUbiomed & 169.50 \\
\hline
Idiap & 178.93 \\
\hline
\textbf{CNNC(128x128, binary)+RBC} & \textbf{224.13} \\
\hline
\textbf{CNNC(96x96, binary)+RBC} & \textbf{237.93} \\
\hline
FEITIJS & 242.46 \\
\hline
Superpixels & 249.34 \\
\hline
VPA SabanciUniv & 261.16 \\
\hline
\textbf{CNNC(96x96, no binarization)+RBC} & \textbf{270.12} \\
\hline
MedGIFT & 317.53 \\
\hline
\end{tabular}
\end{center}
\end{table}

Although it has been shown that in many image classification tasks, using CNN to automatically learn features from images would achieve better performance than human-crafted features, the ability of CNN is mainly limited by the imbalance of training images of the IRMA dataset.

The retrieval time for each image was 1.83 s on average on a laptop with Intel Core i5 CPU and 12 GB RAM. If we enlarge the size of the input images, add more layers or apply more kernels in the CNN, or use more labeled images for training, better results may be achieved. We also tried several new CNN configurations which have deeper architecture and used more kernels, little improvement was obtained. Note that it would take longer to train the CNN and to generate the code, reducing the feasibility of the proposed approach for real-time image retrieval.

We think if we had more training images and enough computational power, the advantages of using deep CNN for medical image retrieval would have been more obvious. In the future, we would create a more balanced dataset to pre-train CNN and average the predictions of several CNNs.

\subsection{ROI Matching Result}
\begin{figure*}[htb]
\centering
\vspace{0.05in}
\centerline{\includegraphics[width=0.9\textwidth]{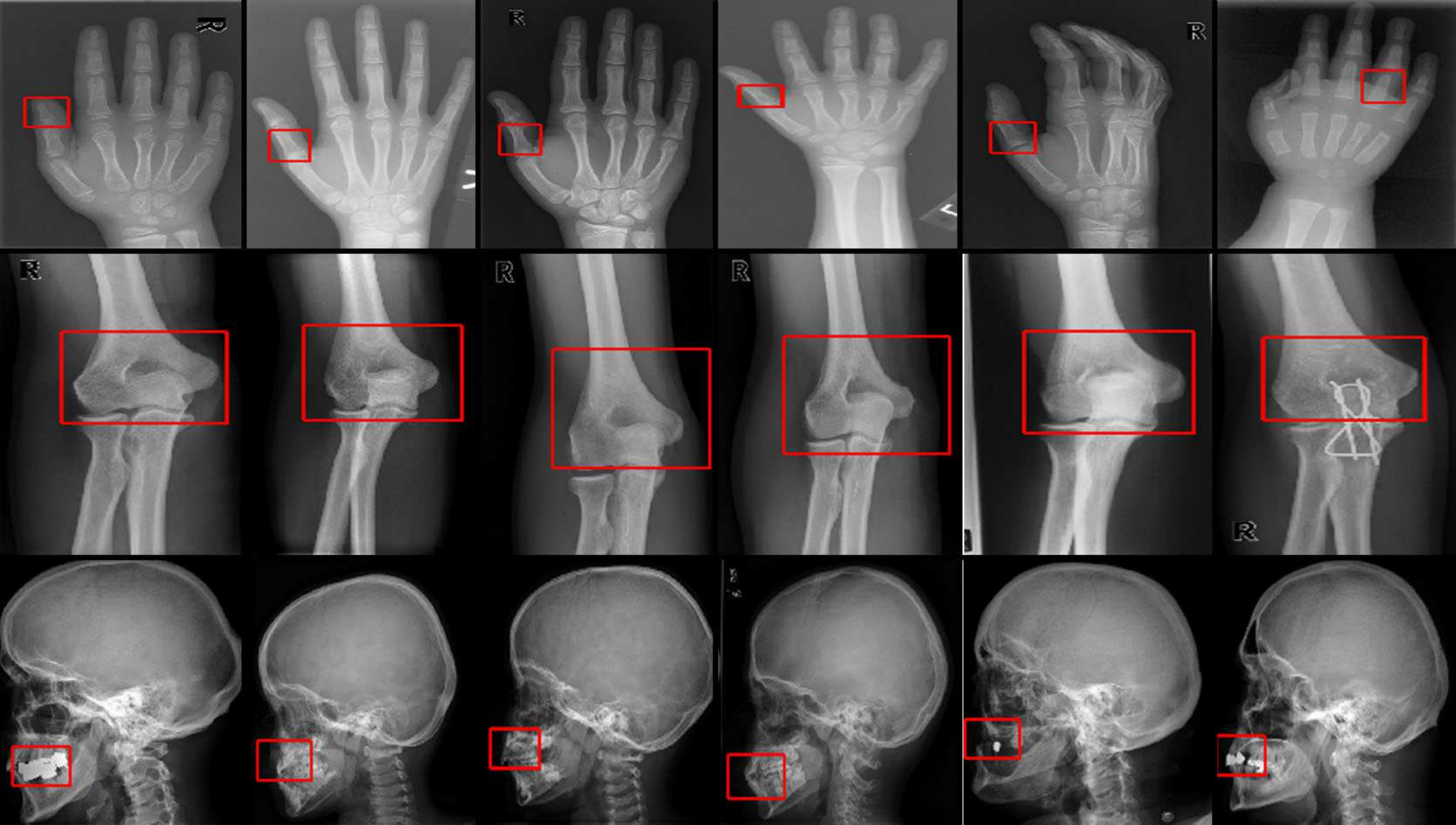}}
\caption{Three examples of ROI matching.}
\label{fig:ROI}
\end{figure*}

ROI matching was further implemented after image retrieval to verify how the search could be extended into finer regions of the image. Note that the ROI in the query image should be selected manually when it is provided by the user. Figure \ref{fig:ROI} shows several examples of ROI matching by using the method described in Section \ref{sec:ROImatch}. The red bounding box in the images in the first column is drawn by the user while the remaining regions are retrieved by ROI matching. However, our approach can find similar sub-regions only based on intensity information, which is not very accurate. Furthermore, the cross-correlation is not rotation and scale invariant. 

\section{Conclusion}
\label{sec:conc}
A simple and fast medical image retrieval method was proposed in this paper based on convolutional neural network and Radon barcodes. Very few feature extraction operations or image pre-processing are needed. The IRMA dataset with 14410 X-ray images was utilized to validate the performance of our method. The error score for our approach was 224.13, better than many approaches published in the CLEF 2009 medical image annotation challenge and later. ROI matching was also investigated based on cross-correlation method. Experimental results suggest that we need publicly available benchmark data for more reliable validations.

\bibliographystyle{IEEEbib}
\bibliography{refs}

\end{document}